\documentclass[11pt]{article}

\usepackage[preprint]{acl}

\usepackage{times}
\usepackage{latexsym}
\usepackage[T1]{fontenc}
\usepackage[utf8]{inputenc}
\usepackage{microtype}

\usepackage{multirow}
\usepackage{array}
\usepackage{booktabs}
\usepackage{amsmath}
\usepackage{amssymb}
\usepackage{amsfonts}
\usepackage{graphicx}
\usepackage{xcolor}
\usepackage{colortbl}
\usepackage{enumitem}

\newcommand{\hi}[1]{\textbf{#1}}
\newcommand{\msurv}{\mathbf{m}}
\newcommand{\ctrunc}{\tilde{c}}

\title{Compression-Aware Abstention: Teaching LLMs\\to Refuse When KV-Compression Masks Remove Answer Evidence}

\author{
  Mohammadali Khodabandehlou \\
  University of Southern California \\
  \texttt{mk40089@usc.edu} \\\And
  Bhaskar Krishnamachari \\
  University of Southern California \\
  \texttt{bkrishna@usc.edu} \\
}

\begin{document}
\maketitle

\begin{abstract}
KV-cache compression reduces LLM inference memory by evicting context tokens, but when the evicted tokens contain answer-bearing evidence, the model may hallucinate instead of recognizing that the compressed context is insufficient. We address this failure from a behavioral perspective: to our knowledge, this is the first work to formulate \emph{compression-aware abstention} as a learning problem, in which a model learns to answer when supporting evidence survives compression and abstain when it does not. We construct supervision from compressor survival masks and tight answer-bearing spans, labeling examples as \textsc{Confident} when evidence survives and \textsc{Abstain} when it is removed. A 10.1M-parameter LoRA adapter trained on $\sim$2.6K MuSiQue 2-hop QA examples reduces base-model hallucinations by 97\% under prompt-style truncation while preserving correct answering on evidence-retaining examples. Unlike prompt-only abstention baselines, which over-abstain on many answerable high-retention examples, the trained adapter learns a conditional policy. We also evaluate the method under actual compressed-cache decoding, where multi-compressor training yields a 6--22$\times$ relative lift over the unaided base on evidence-retaining examples. Controlled-deletion experiments show that the learned behavior is driven by evidence content rather than input length alone.
\end{abstract}

\section{Introduction}
\label{sec:intro}

KV-cache compression methods reduce inference memory by evicting context tokens deemed unimportant by some scoring rule \cite{li2024snapkv,zhang2023h2o,kim2025kvzip,devoto2025expectedattention}. They are designed to be \emph{behavior-preserving}, on the assumption that low-importance tokens carry no answer-relevant signal. When the eviction rule drops a token that is answer-bearing for some downstream query, the model is left with an input missing exactly the evidence it needs, and tends to fabricate a confident wrong answer rather than acknowledge the loss. The qualitative pattern is documented \cite{yang2024mikv}; existing remedies are representation-side (keep evicted entries in low-precision form rather than dropping them).

We propose a behavioral remedy: train the model to recognize when the surviving context cannot support an answer and respond honestly, answering when the supporting evidence survives and refusing when it does not. We refer to this as \emph{compression-aware abstention}. To our knowledge, this is the first work to address KV-cache-compression-induced evidence loss from a behavioral perspective, by training the model to abstain when the surviving compressed context no longer supports an answer.

\paragraph{Scope.} We construct training data from compression masks and feed the surviving tokens to the model as a normal prompt; we then test, separately, whether the trained behavior survives when the model decodes from a physically compressed cache (KVzipPress at inference). \S\ref{sec:cctraining} reports both inference paths: the prompt-style (ps) path used to build training data and the compressed-cache (cc) path that matches deployment. Our objective is complementary to training-aware compression such as KV-Distill \cite{chari2025kvdistill}, which instead preserves uncompressed-model behavior under compression (\S\ref{sec:related}).

\paragraph{Contributions.} We make five contributions:

\noindent\textbf{C1.} We formulate compression-aware abstention as a behavioral learning problem. Rather than only asking whether KV-cache compression preserves the uncompressed model's answer, we ask whether the compressed context still contains enough evidence to support an answer. We define examples as \textsc{Confident} when answer-bearing evidence survives compression and \textsc{Abstain} when it does not, and train the model to answer or refuse accordingly.

\noindent\textbf{C2.} We show that tight evidence spans are crucial for making this learning problem well-posed. Paragraph-level support labels wash out the effect of compression because long support paragraphs often remain partially present even when key evidence is lost. By labeling survival over short answer-bearing spans, we obtain the per-example variation needed to distinguish answerable from unanswerable compressed contexts.

\noindent\textbf{C3.} We show that a small adapter can sharply reduce compression-induced hallucination while preserving answerability. A 10.1M-parameter LoRA adapter trained on approximately 2.6K compression-labeled MuSiQue examples eliminates 97\% of base-model hallucinations under prompt-style truncation, while preserving correct answering on \textsc{Confident}-gold examples.

\noindent\textbf{C4.} We show that learned abstention is more conditional than prompt-only abstention and more robust when trained across compressors. Prompt-only baselines tend to over-abstain when evidence is actually present: at high retention, they abstain on about half of \textsc{Confident}-gold examples, while the trained adapter abstains on only 6--10\%. Training on multiple compressor masks further improves transfer beyond a single compression pattern.

\noindent\textbf{C5.} We evaluate the method under deployment-relevant compressed-cache decoding and test what the adapter has learned. Under actual compressed-cache inference, the unaided base model nearly collapses on evidence-retaining examples; multi-compressor training improves correct answering from 4\% to 25\% aggregate and from 1.4\% to 32\% at high retention. Controlled-deletion experiments show that the adapter's decisions depend on which evidence tokens survive, not merely on input length.

\section{Compression-aware abstention as a learning problem}
\label{sec:formalism}

We give a formal definition of the learning problem the rest of the paper addresses. The definition makes three things precise: (i) what a compression-truncated input is, (ii) how the gold label is determined from the survival mask and the supporting-span structure, and (iii) what the trained adapter optimizes.

\paragraph{A motivating example.} Consider the 2-hop question \emph{``Who founded the company that distributed the film UHF?''}, whose answer is \emph{Mike Medavoy}. Deriving this answer from the context requires one short, specific piece of evidence (the phrase naming the founder) to survive compression. Now suppose a KV-cache compressor keeps most of the surrounding context (the film, its release, other people and companies) but, scoring that particular name as low-importance, evicts the few tokens that spell it out. The truncated context still reads fluently and still looks on-topic, so a base model prompted with it tends to supply a confident but unsupported name (in our data the base answers ``Kevin Smith''). The correct behavior depends entirely on whether that one phrase survived: if the founder's name is kept, the example is \textsc{Confident} and the model should answer \emph{Mike Medavoy}; if it is evicted, the example is \textsc{Abstain} and the model should refuse, because the surviving text no longer determines the answer. The rest of this section makes ``a supporting phrase survives'' quantitative (Eq.~\ref{eq:rho}) and turns this all-or-nothing flip into a training label (Eq.~\ref{eq:label}).

\paragraph{Compression-truncated input.}
Let $c \in \mathcal{V}^{n_c}$ be a context tokenized into $n_c$ tokens, and let $q$ be a query. A KV cache compressor at retention ratio $r$ produces a per-token \emph{survival mask} $\msurv \in \{0,1\}^{n_c}$ with $\tfrac{1}{n_c}\sum_i m_i \approx r$, and the associated compression-truncated input $\ctrunc(\msurv) = (c_i : m_i = 1)$ concatenates the surviving tokens. Different compressors select \emph{which} tokens survive; the model is then run on $\ctrunc$ as a standard prompt.

\paragraph{Evidence survival.}
Let $S = \{S_1, \ldots, S_K\}$ be a set of \emph{supporting spans}: contiguous token ranges in $c$ that are jointly required for the answer $a$ to be derivable from the context. Define the \emph{evidence-survival fraction} of span $S_k$ under mask $\msurv$ as
\begin{equation}
\rho(\msurv, S_k) = \frac{\sum_{i \in S_k} m_i}{|S_k|}.
\label{eq:rho}
\end{equation}
The granularity of $S_k$ is decisive (\S\ref{sec:tightspan}): when $|S_k|$ is large, $\rho$ averages out and per-example label variance vanishes; when $|S_k|$ is the size of the verbatim answer string (median 3 tokens), $\rho$ is informative.

\paragraph{Gold label.}
Given thresholds $\tau_{\mathrm{low}} < \tau_{\mathrm{high}} \in [0, 1]$, the gold label $y(\msurv, S)$ is
\begin{equation}
\label{eq:label}
y \!=\!
\begin{cases}
\textsc{Confident} & \!\!\text{if } \min_k \rho(\msurv, S_k) \!\geq\! \tau_{\mathrm{high}} \\
\textsc{Abstain} & \!\!\text{if } \min_k \rho(\msurv, S_k) \!<\! \tau_{\mathrm{low}} \\
\textsc{Drop} & \!\!\text{otherwise (excluded).}
\end{cases}
\end{equation}
Examples whose minimum span survival lands in the uncertain zone $[\tau_{\mathrm{low}}, \tau_{\mathrm{high}})$ are dropped from training to avoid noisy targets.

\paragraph{Target string and learning objective.}
Each $(c, q, a, \msurv, S)$ tuple is mapped to a target string $t = a$ if $y = \textsc{Confident}$ and $t = t_\bot$ if $y = \textsc{Abstain}$, where $t_\bot$ is a fixed refusal string. The training set $\mathcal{D}$ consists of $(\msurv, q, t)$ triples for which $y \neq \textsc{Drop}$, with truncated input $\ctrunc(\msurv)$. The adapter parameters $\theta$ are LoRA weights \cite{hu2022lora} attached to a frozen base LLM $f$, optimizing
\begin{equation}
\label{eq:loss}
\mathcal{L}(\theta) = -\!\!\!\!\!\sum_{(\msurv, q, t) \in \mathcal{D}}\!\!\!\! \log f_\theta\bigl(t \mid \ctrunc(\msurv),\, q\bigr).
\end{equation}
This loss splits over the two label classes, $\mathcal{L}(\theta) = \mathcal{L}_{\textsc{conf}}(\theta) + \mathcal{L}_{\textsc{abs}}(\theta)$, summing the cross-entropy over $\mathcal{D}_{\textsc{conf}}$ (target $a$) and $\mathcal{D}_{\textsc{abs}}$ (target $t_\bot$). The two terms incentivize opposite per-example behaviors: $\mathcal{L}_{\textsc{conf}}$ penalizes refusing when the truncated input does support the answer, $\mathcal{L}_{\textsc{abs}}$ penalizes answering when it does not. Both are needed for calibrated bidirectional behavior: optimizing only $\mathcal{L}_{\textsc{abs}}$ yields a fixed abstention prior (the prompt-only failure mode of \S\ref{sec:baselines}), while optimizing only $\mathcal{L}_{\textsc{conf}}$ approaches a preservation-style objective \citep{chari2025kvdistill} (always answer). We offer this as the natural reading of Eq.~\ref{eq:loss}, not as an ablated claim.

\paragraph{Class balance and comparison.}
Because the label distribution is U-shaped in $r$ (mostly \textsc{Abstain} at low retention, mostly \textsc{Confident} at high), the mid-retention regime ($r \approx 0.50$) is where both classes co-occur and Eq.~\ref{eq:label} is informative per-example; we apply a per-(compressor, $r$) class-balance floor at $r \geq 0.20$ so both loss terms see comparable counts (\S\ref{sec:training}, Appendix~\ref{app:mixture}). The objective is complementary to preservation-style training \cite{chari2025kvdistill}: once $\rho(\msurv, S) < \tau_{\mathrm{low}}$, the uncompressed-model answer is no longer a meaningful target, and the right target is $t_\bot$.

\section{Method}
\label{sec:method}

\subsection{Tight-span localization}
\label{sec:tightspan}

Eq.~\ref{eq:rho} is informative only when $|S_k|$ is small. For each MuSiQue \cite{trivedi2022musique} 2-hop example, the \texttt{question\_decomposition} field gives each single-hop sub-answer as a short atomic string (e.g.,\ ``Mike Medavoy''); we locate each verbatim in its supporting paragraph, yielding a short span (median 3 tokens, 99th percentile 11; located in 100\% of our pool).

Tight spans are what make the labeling discriminative. With full supporting paragraphs ($\sim$100 tokens) as $S_k$, $\rho$ averages out: at $r = 0.50$ every paragraph keeps roughly half its tokens, so every example falls in the uncertain zone and is dropped (label entropy $0.000$, vs $0.95$ for tight spans; Appendix~\ref{app:label_entropy}). The pivot to verbatim answer spans is the reason the rest of the paper has signal to fit.

\subsection{Pipeline and adapter}
\label{sec:training}

\begin{figure}[t]
\centering
\includegraphics[width=\columnwidth]{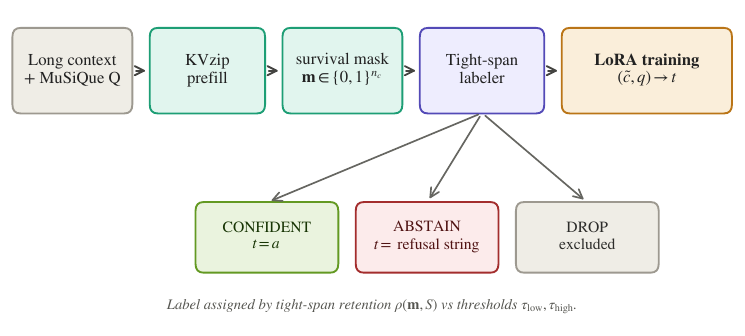}
\caption{Training pipeline. KVzip prefill produces a per-token survival mask $\msurv$ at the target retention ratio. The labeler computes evidence-survival $\rho(\msurv, S_k)$ (Eq.~\ref{eq:rho}) over tight supporting spans and assigns \textsc{Confident}, \textsc{Abstain}, or \textsc{Drop} via Eq.~\ref{eq:label}. The LoRA adapter is trained on (truncated input $\ctrunc$, query $q$) $\to$ target $t$.}
\label{fig:pipeline}
\end{figure}

Figure~\ref{fig:pipeline} summarizes the construction pipeline. The compressor produces $\msurv$; the labeler computes $\rho(\msurv, S_k)$ for each tight span and assigns $y$ via Eq.~\ref{eq:label} with thresholds $\tau_{\mathrm{low}} = 0.3$, $\tau_{\mathrm{high}} = 0.8$ (set a priori). The resulting (truncated input, query, target) triple enters $\mathcal{D}$ if $y \neq \textsc{Drop}$ (per-ratio label entropy in Appendix~\ref{app:label_entropy}).

We attach a 10.1M-parameter LoRA adapter \cite{hu2022lora} (rank 16) to the four attention projections (\texttt{q\_proj}, \texttt{k\_proj}, \texttt{v\_proj}, \texttt{o\_proj}; MLP modules unadapted), $0.13\%$ of Qwen2.5-7B-Instruct's parameters. Optimizer, learning-rate sweep, and precision settings are in Appendix~\ref{app:trainingdetails}.


Train $\approx$2.6K instances (30/70 minority-class floor); validation 463 at the natural class distribution, disjoint by example ID. We call this KVzip-only adapter \emph{v1}; \S\ref{sec:crosscompressor} introduces a multi-compressor variant, \emph{mix}, and \S\ref{sec:cctraining} distinguishes the prompt-style and compressed-cache versions of each (v1-ps, mix-ps, and mix-cc).

\section{Experimental setup}
\label{sec:setup}

\paragraph{Models.} Qwen2.5-7B-Instruct \cite{qwen2025qwen25} for all main results; Llama-3.1-8B-Instruct \cite{grattafiori2024llama3} as a held-out base model.

\paragraph{Compressors.} KVzip \cite{kim2025kvzip} for training masks, via the authors' reference implementation: kvpress's \texttt{KVzipPress} defers eviction in a way incompatible with prompt-style mask extraction, though \S\ref{sec:cctraining} uses it for compressed-cache inference. Expected Attention \cite{devoto2025expectedattention} and SnapKV (context-only and native query-aware modes) \cite{li2024snapkv} are held-out compressors, accessed via the kvpress library \cite{kvpress2025}. All are calibrated to retain an identical number of context tokens at each $r$.

\paragraph{Retention ratios.} $r \in \{0.05,\allowbreak 0.10,\allowbreak 0.20,\allowbreak 0.30,\allowbreak 0.50,\allowbreak 0.80\}$. The gold class distribution is U-shaped in $r$ (Table~\ref{tab:headline}): almost all \textsc{Abstain} at low retention, almost all \textsc{Confident} at high, and balanced at the headline mid-slice $r = 0.50$.

\paragraph{Outputs and metric.} Greedy decoding, max 64 new tokens. Outputs are classified into six categories (Appendix~\ref{app:eval_categories}). \emph{Honest accuracy} is the fraction of outcomes that are either \textsc{Confident}-gold with the model emitting the answer $a$, or \textsc{Abstain}-gold with the model emitting the refusal $t_\bot$.

\paragraph{Compute and reproducibility.} All training and evaluation run on a single NVIDIA H100 80GB GPU. Code, evaluation outputs, adapters, datasets, and a script that recomputes every reported number are released.\footnote{\url{https://github.com/mali-kh/compression-aware-abstention}}

\section{Results}
\label{sec:results}

\subsection{Headline}
\label{sec:headline}

\begin{table}[t]
\centering
\small
\setlength{\tabcolsep}{4.0pt}
\begin{tabular}{cccccc}
\toprule
$r$ & $n$ & C/A & base & adp & $\Delta$hon \\
\midrule
0.05 & 100 & 0/100 & 0.15 & 1.00 & $+0.85$ \\
0.10 & 95  & 1/94  & 0.27 & 0.99 & $+0.72$ \\
0.20 & 82  & 2/80  & 0.30 & 0.92 & $+0.62$ \\
0.30 & 65  & 3/62  & 0.35 & 0.79 & $+0.44$ \\
\rowcolor{yellow!18}
\hi{0.50} & \hi{48} & \hi{22/26} & \hi{0.44} & \hi{0.66} & \hi{$+0.22$} \\
0.80 & 73 & 69/4 & 0.51 & 0.60 & $+0.09$ \\
\midrule
\textit{agg.} & 463 & 97/366 & 0.32 & 0.86 & $+0.54$ \\
\bottomrule
\end{tabular}
\caption{Per-ratio honest accuracy on Qwen val. Mean over three seeds (42, 43, 44). The 0.50 row is the only ratio at which the gold class is balanced and at which a model conditioning on $\ctrunc$ can do better than a fixed prior (\S\ref{sec:baselines}).}
\label{tab:headline}
\end{table}

\begin{figure}[t]
\centering
\includegraphics[width=\columnwidth]{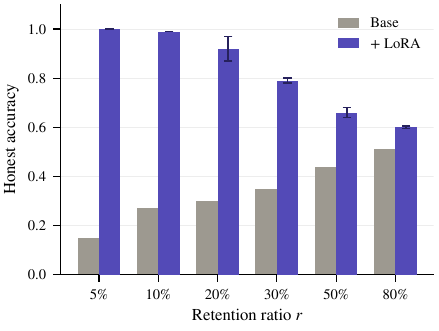}
\caption{Per-ratio honest accuracy on Qwen val (KVzip masks). Bars are means across three seeds; error bars are seed spread. The adapter improves honest accuracy at every retention ratio.}
\label{fig:perratio}
\end{figure}

Training improves honest accuracy at every retention ratio (Table~\ref{tab:headline}, Figure~\ref{fig:perratio}), and the gain is driven by a collapse in hallucination: confident wrong answers on \textsc{Abstain}-gold (the \emph{hallucinated} category of Appendix~\ref{app:eval_categories}) fall from 232 to 7. The adapter's abstention rate falls monotonically as retention rises, consistent with evidence-sensitive behavior, though this alone does not rule out a retention-ratio heuristic, which \S\ref{sec:baselines} and \S\ref{sec:mechanism} test directly. Conditional answer quality is preserved (Appendix~\ref{app:calibration}).

\paragraph{No over-abstention cost.} A refusal-biased adapter could cut hallucinations on \textsc{Abstain}-gold while also refusing answerable \textsc{Confident}-gold examples. It does not: the adapter's correct-answer rate on \textsc{Confident}-gold is at least the base's in all eight prompt-style cells across four compressors (\emph{answerability preservation} ratio $1.00$--$1.36$; Appendix~\ref{app:answerability}).

\subsection{Baselines}
\label{sec:baselines}

Aggregate honest accuracy is useful for orientation, but it is not a reliable
operationalization of compression-aware abstention by itself. The reason is that
the gold labels are highly skewed as a function of the retention ratio $r$: at
low $r$, almost all examples are \textsc{Abstain}-gold, while at high $r$, almost
all are \textsc{Confident}-gold. A trivial policy that never sees the truncated
input $\ctrunc$ can exploit this structure: predict \textsc{Abstain} when
$r < \tau$ and \textsc{Confident} otherwise. Sweeping $\tau$ on Qwen val, the
best threshold ($\tau = 0.80$) reaches aggregate honest accuracy $0.93$,
exceeding both v1 ($0.84$) and the mixture adapter ($0.86$). This does not make
the threshold policy compression-aware; it only shows that aggregate accuracy
rewards the U-shaped class distribution. We therefore use the diagnostic slices
below to ask whether a method is actually sensitive to the evidence that survives
in $\ctrunc$.

\paragraph{Two diagnostic slices.}
The balanced mid-retention slice and the high-retention slice probe complementary
failure modes. At $r = 0.50$, the gold labels are nearly balanced
(22 \textsc{Confident}, 26 \textsc{Abstain}), so a retention-ratio threshold is
forced into one fixed prediction across the slice. This slice tests whether a
method can distinguish answerable from unanswerable truncated inputs on a
per-example basis. At $r = 0.80$, the gold labels are dominantly
\textsc{Confident}; this slice tests the opposite failure mode, namely whether an
abstention method over-refuses even when evidence usually survives. Together,
these slices separate genuine evidence sensitivity from fixed abstention or
fixed-answering priors.

\paragraph{Baseline definitions.}
\emph{P2 prompt}: a light-touch abstention instruction prepended to the user
message:
``\textit{If the available context does not contain enough evidence to answer the
question, respond with: `I cannot determine the answer to this question from the
available context.' Otherwise, answer the question.}''
No LoRA adapter is applied.
\emph{Fragment}: the same compression-truncated input $\ctrunc$ is presented to a
clean base model, with no LoRA adapter and no abstention instruction. The
Fragment baseline is input-matched to our adapter; the only difference is the
learned LoRA weights.
\emph{best-$\tau$}: the best ratio-only threshold policy, with $\tau = 0.80$
tuned on KVzip aggregate.

\begin{table}[t]
\centering
\small
\setlength{\tabcolsep}{3.4pt}
\begin{tabular}{lcccccc}
\toprule
Compressor & gold (C/A) & best-$\tau$ & P2 & Frag & v1 & Mix \\
\midrule
KVzip          & 22/26 & 0.54 & 0.54 & 0.44 & 0.65 & \hi{0.69} \\
ExpAttn        & 23/19 & 0.55 & 0.60 & 0.38 & 0.67 & \hi{0.71} \\
SnapKV (ctx)   & 18/21 & 0.54 & 0.64 & 0.33 & 0.46 & \hi{0.64} \\
\textit{SnapKV (nat)} & \textit{28/9} & \textit{0.76} & \textit{0.54} & \textit{0.38} & \textit{0.41} & \textit{0.51} \\
\bottomrule
\end{tabular}
\caption{Honest accuracy at $r = 0.50$ across compressors. \emph{v1}: adapter trained on KVzip masks only. \emph{Mix}: adapter trained on KVzip $+$ ExpAttn $+$ SnapKV masks jointly. SnapKV-native at this slice has degenerate gold mix (28/9) and is not interpretable for this comparison; reported in italics for completeness.}
\label{tab:baselines050}
\end{table}

\paragraph{Balanced slice: training helps; prompt inconclusive.}
Table~\ref{tab:baselines050} shows that the mixture adapter improves over the
Fragment baseline by 25--33pp at $r = 0.50$ on the three interpretable
compressor settings (per-compressor paired McNemar exact,
$p \in [0.0005, 0.012]$ on KVzip / ExpAttn / SnapKV-context; not significant on
SnapKV-native). This establishes that the behavior is not obtained merely by
showing the base model the surviving text.

The comparison to the prompt-only baseline is more nuanced at this slice. Point
estimates favor the mixture adapter on KVzip ($+0.15$) and ExpAttn ($+0.12$),
tie on SnapKV-context ($\pm 0$), and slightly favor P2 on SnapKV-native
($-0.03$); none reaches per-compressor significance at
$n \in [37,48]$ (per-compressor exact $p$-values and paired-bootstrap CIs in Appendix~\ref{app:baseline_stats}).
Pooling the four compressors into a 166-example paired test yields
$\Delta = +0.07$ ($p = 0.17$). Thus the mid-slice suggests a positive trend over
prompting, but the available sample does not make that comparison decisive. The
cleaner prompt-vs-training result appears at high retention, where the same
fixed-prior failure has a larger and more measurable effect.

\begin{figure}[t]
\centering
\includegraphics[width=\columnwidth]{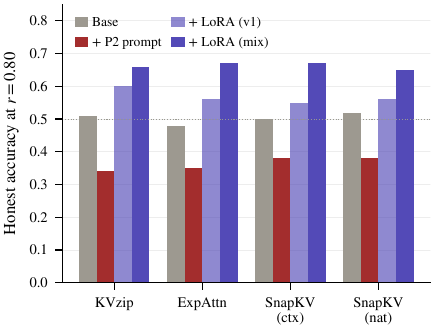}
\caption{Honest accuracy at $r = 0.80$ (high retention, dominantly \textsc{Confident}-gold) across four compressors. The P2 abstention prompt drops below base on every compressor (dotted line at base level); the trained adapter does not. Calibrated bidirectional behavior (abstain when evidence is missing \emph{and} answer when it survives) requires per-example sensitivity to $\ctrunc$, which the loss decomposition in Eq.~\ref{eq:loss} explicitly captures (\S\ref{sec:formalism}).}
\label{fig:highfill}
\end{figure}

\paragraph{Fixed prior vs.\ learned policy.}
At $r = 0.80$, where roughly 95\% of the gold labels are \textsc{Confident},
the prompt-only baseline exposes its main failure mode: it over-abstains even
when evidence usually survives. On KVzip, P2 honest accuracy drops to $0.34$,
below the base model's $0.51$. It abstains on $49\%$ of the 69
\textsc{Confident}-gold examples, compared with $10\%$ for the mixture adapter
and $3\%$ for v1. The pattern is uniform across compressors
(Figure~\ref{fig:highfill}): P2 over-abstains on $49$--$51\%$ of
\textsc{Confident}-gold examples, while the mixture adapter over-abstains on only
$6$--$10\%$.

Restricting to the \textsc{Confident}-gold subset at $r = 0.80$, the mixture
adapter outperforms P2 on every compressor, with
$\Delta \in [+0.28,+0.34]$ and per-compressor paired McNemar exact
$p \leq 2 \times 10^{-4}$. Pooled across the 260 \textsc{Confident}-gold
examples at this retention, the advantage is $\Delta = +0.32$
(95\% paired-bootstrap CI $[+0.25,+0.38]$, McNemar
$p \approx 1.5 \times 10^{-16}$). This is the clearest baseline evidence in our
setting: an abstention prompt tends to create a broad refusal prior, whereas the
trained adapter learns a bidirectional policy that answers when evidence survives
and abstains when it does not. This is precisely the behavior encouraged by the
loss decomposition in Eq.~\ref{eq:loss} into
$\mathcal{L}_{\textsc{conf}}$ and $\mathcal{L}_{\textsc{abs}}$. Whether richer
prompting strategies (for example, self-consistency over abstention judgments,
chain-of-thought introspection, or verifier-style confidence checks) could
approximate this behavior at inference time remains open; we do not test them
here.

\paragraph{Surviving text is not enough.}
Finally, the Fragment baseline shows that correct abstention is a learned
property of the adapter, not an automatic consequence of seeing a shortened
context. Of the 89 examples where v1 correctly abstained on KVzip but
hallucinated on SnapKV (the cross-compressor failure mode of
\S\ref{sec:crosscompressor}), the Fragment baseline catches only
$13/89 = 15\%$ on the KVzip side: it hallucinates on 80\% of the same examples
the adapter abstains on.

\subsection{Cross-compressor transfer and mixture training}
\label{sec:crosscompressor}

\begin{figure}[t]
\centering
\includegraphics[width=\columnwidth]{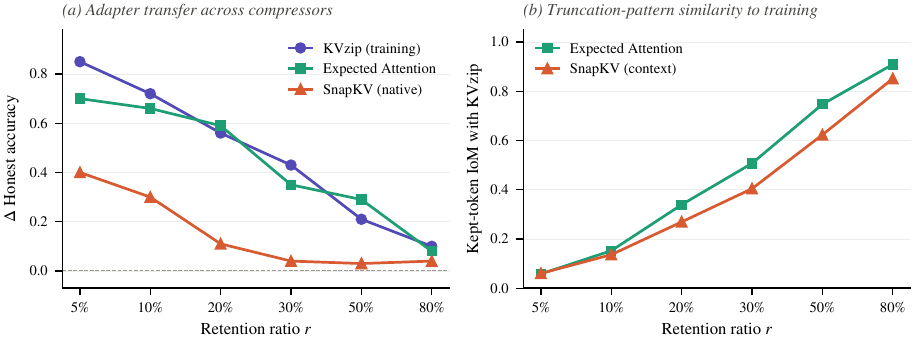}
\caption{(a) Per-ratio $\Delta$honest accuracy of the v1 adapter (KVzip-trained) on three held-out compressors. Within-family transfer is clean (KVzip $\to$ ExpAttn, aggregate $+0.49$ vs $+0.53$); cross-family transfer is partial (KVzip $\to$ SnapKV-native, $+0.20$). (b) Differential transfer tracks kept-token intersection-over-min between each compressor and KVzip.}
\label{fig:crosscomp}
\end{figure}

\begin{table}[t]
\centering
\small
\begin{tabular}{lcccc}
\toprule
Compressor & v1 & Mix & mix $-$ v1 & Rescue \\
\midrule
KVzip                & $+0.53$ & $+0.55$ & $+0.02$ & --- \\
Expected Attn        & $+0.49$ & $+0.57$ & $+0.08$ & --- \\
\rowcolor{yellow!18}
\hi{SnapKV (ctx)}    & \hi{$+0.21$} & \hi{$+0.53$} & \hi{$+0.32$} & \hi{82/89} \\
\rowcolor{yellow!18}
\hi{SnapKV (nat)}    & \hi{$+0.20$} & \hi{$+0.43$} & \hi{$+0.23$} & --- \\
\bottomrule
\end{tabular}
\caption{Aggregate $\Delta$honest accuracy: v1 (KVzip-only) vs mixture training (KVzip $+$ ExpAttn $+$ SnapKV-context). \emph{v1}: seed 42; \emph{Mix}: mean across seeds 42 and 43. \emph{Rescue}: of 89 v1-KVzip-correct $\to$ v1-SnapKV-hallucinated examples, fraction the mixture catches under SnapKV evaluation (seed 42). Per-cell variance in Appendix~\ref{app:mixture}.}
\label{tab:mixture}
\end{table}

A compression-aware abstention policy should not depend too tightly on the
idiosyncrasies of one compressor's mask. To test this, we first train v1 only on
KVzip masks and evaluate it, without retraining, on three held-out compressors.
The result is mixed. Transfer to Expected Attention is strong: the aggregate
gain remains close to the KVzip setting ($+0.49$ vs $+0.53$). Transfer to SnapKV
is much weaker, with aggregate gains of only $+0.21$ for SnapKV-context and
$+0.20$ for SnapKV-native (Figure~\ref{fig:crosscomp}a).

This pattern suggests that the adapter has learned two different behaviors. One
is compressor-agnostic: when the surviving text contains enough evidence, the
model can read and answer. The other is more compressor-specific: recognizing
that evidence is missing depends partly on the deletion patterns seen during
training. The transfer results track this interpretation. At $r = 0.50$, KVzip
and Expected Attention agree on $75\%$ of kept tokens, while KVzip and SnapKV
agree on only $62\%$; the drop in transfer follows the same ordering
(Figure~\ref{fig:crosscomp}b). Appendix~\ref{app:crosscompressor} gives a
qualitative decomposition of these ``read-and-answer'' and ``recognize-missing-evidence''
behaviors.

We therefore retrain on a mixture of masks from KVzip, Expected Attention, and
SnapKV-context. The mixture set contains 8772 training instances, with
per-(compressor, $r$) class balance for $r \geq 0.20$ and the natural class
distribution at the extreme low-retention ratios. Hyperparameters match v1
except for $\mathrm{lr}=2{\times}10^{-4}$; results are averaged over two seeds.

Mixture training largely closes the cross-compressor gap without sacrificing
performance on the original compressor (Table~\ref{tab:mixture}). On KVzip, the
aggregate gain is essentially unchanged ($+0.53 \to +0.55$). On Expected
Attention, it improves from $+0.49$ to $+0.57$. The largest gains appear on
SnapKV, where v1's partial transfer is the main failure mode: SnapKV-context
improves from $+0.21$ to $+0.53$, and SnapKV-native from $+0.20$ to $+0.43$.
The rescue analysis makes the same point at the example level: of the 89 cases
where v1 abstained correctly under KVzip but hallucinated under SnapKV, mixture
training catches 82 under SnapKV evaluation.

Mixture training also improves calibration. The conditional wrong-when-answering
rate decreases on 7 of 8 conclusive cells (Appendix~\ref{app:calibration}), and
the KVzip-specific overcaution seen in v1 at $r = 0.50$ disappears: mixture
overcaution is $23\%$, matching the base model. Thus the benefit of mixture
training is not merely higher honest accuracy; it makes the learned abstention
policy less tied to a single compressor's mask geometry.

\subsection{What is the load-bearing training signal?}
\label{sec:randommask}

Retraining v1 with masks drawn uniformly at random (at matched per-example retention counts) matches the mixture adapter on KVzip val and beats KVzip-only v1 on cross-family transfer. Under prompt-style inference, the load-bearing signal is therefore the abstention objective itself, not the importance-weighted structure of the compressor mask; single-compressor training additionally induces an anti-transfer signature. This motivates compressor-mixture training (\S\ref{sec:crosscompressor}), which \S\ref{sec:cctraining} extends to compressed-cache inference. Full numbers are in Appendix~\ref{app:randommask}.

\subsection{Compressed-cache inference and cc-training}
\label{sec:cctraining}

The results so far use prompt-style inference: we apply the compressor's survival
mask, feed the surviving tokens back to the model as an ordinary prompt, and
measure whether the adapter answers or abstains. This is useful for controlled
training and analysis, but it is not the deployment path. In actual
compressed-cache inference, the model first prefills on the full context, the
compressor evicts low-scoring K/V pairs from the cache, and generation proceeds
from the compressed cache state. We therefore ask whether the learned abstention
behavior survives this more realistic setting. We evaluate on the same Qwen
validation set using kvpress's KVzipPress at inference.

\begin{table}[t]
\centering
\small
\setlength{\tabcolsep}{3.4pt}
\begin{tabular}{lcccc}
\toprule
$r$ & base-cc & v1-ps cc & mix-ps cc & mix-cc cc \\
\midrule
0.05 & --- & 0.560 & 1.000 & \hi{1.000} \\
0.10 & --- & 0.495 & 0.958 & \hi{0.989} \\
0.20 & --- & 0.415 & 0.829 & \hi{0.963} \\
0.30 & --- & 0.323 & 0.600 & \hi{0.923} \\
\rowcolor{yellow!18}
\hi{0.50} & --- & \hi{0.271} & \hi{0.354} & \hi{0.500} \\
0.80 & --- & 0.233 & 0.315 & 0.315 \\
\midrule
\textit{agg.} & $0.186$ & $0.406$ & $0.730$ & \hi{$0.821$} \\
\bottomrule
\end{tabular}
\caption{Per-ratio honest accuracy under \emph{actual} compressed-cache inference (KVzipPress) on Qwen val ($n = 463$). \emph{base-cc}: unmodified Qwen2.5-7B-Instruct under cc inference (per-ratio omitted; aggregate $0.186$). \emph{v1-ps}: prompt-style-trained, KVzip-only. \emph{mix-ps}: prompt-style-trained, multi-compressor. \emph{mix-cc}: compressed-cache-trained, multi-compressor (seed 42; per-seed values and two-seed mean in Appendix~\ref{app:cctraining}).}
\label{tab:cctraining}
\end{table}

Compressed-cache decoding is substantially harder for the unaided base model.
Under KVzipPress, base-cc reaches aggregate honest accuracy $0.186$, which is
13pp below its prompt-style counterpart. The collapse is most visible on
\textsc{Confident}-gold examples, where the correct-answer rate is only
$4/97$ aggregate and $1/69$ at $r = 0.80$. In other words, when the answer-bearing
evidence is labeled as surviving, the instruction-tuned base model still often
fails to use it once generation proceeds from the compressed cache.

The trained adapters recover much of this loss. Prompt-style multi-compressor
training transfers surprisingly well to compressed-cache inference: mix-ps
reaches aggregate honest accuracy $0.730$, a $+0.544$ gain over base-cc. By
contrast, the KVzip-only prompt-style adapter transfers much less well:
v1-ps reaches $0.406$, only $+0.220$ over base-cc. This mirrors the
cross-compressor pattern in \S\ref{sec:crosscompressor}: training on a single
mask family leaves the adapter more brittle when the inference-time corruption
pattern changes.

Training directly in the compressed-cache regime adds another gain. Using the
same multi-compressor mixture but computing the loss after cache eviction,
mix-cc reaches aggregate honest accuracy $0.821$, improving by $+0.091$ over
mix-ps under the same cc inference path. The largest gains are at the intermediate
retention ratios, especially $r = 0.30$ and $r = 0.50$, where mix-cc improves
over mix-ps by $+0.32$ and $+0.15$, respectively. At $r = 0.80$, mix-cc and
mix-ps tie in honest accuracy ($0.315$ vs $0.315$; McNemar on
\textsc{Confident}-gold, $p = 0.80$), indicating that high-retention
compressed-cache answering remains the hardest case.

The answerability numbers clarify the scale of the recovery. On
\textsc{Confident}-gold examples, multi-compressor cc-training improves correct
answering from $4/97$ to $24/97$ aggregate, and from $1/69$ to $22/69$ at
$r = 0.80$. This corresponds to a $6$--$22\times$ relative lift over base-cc,
with non-overlapping Wilson 95\% confidence intervals
(Appendix~\ref{app:answerability}). The gain is large because base-cc nearly
collapses in this regime; the absolute rates also show that cc inference remains
more difficult than prompt-style truncation.

Multi-compressor training is important even when training directly under
compressed-cache decoding. A single-compressor cc-trained adapter ties mix-cc on
aggregate honest accuracy, but at $r = 0.80$ it produces a $61\%$
wrong-answer rate on \textsc{Confident}-gold examples, compared with $30\%$ for
mix-cc (Appendix~\ref{app:cctraining}). Thus the mixture is not merely improving
the aggregate score; it keeps the answer-vs-abstain tradeoff better calibrated
under the deployment-relevant inference path.

\subsection{Mechanism: evidence content or input length?}
\label{sec:mechanism}

The adapter's decisions track \emph{which} evidence survives, not how much: it responds to evidence content, not input length. This matters because the adapter sees a shortened input and could in principle be abstaining on length alone. Three controlled-deletion tests, all holding input length fixed, rule that out (Appendix~\ref{app:mechanism}). The decisive one constructs, for each \textsc{Confident}-gold example at $r = 0.50$, two \emph{equal-length} masks: one keeping KVzip's highest-priority half, one its lowest-priority half. The adapter answers on the high-priority masks and abstains on the low-priority ones, where neither adapter answers a single example correctly and mix-ps abstains on all $22$ with \emph{zero} confident wrong answers; replacing only the answer-bearing tokens with random kept tokens raises abstention just as sharply. A length prior predicts identical behavior across these matched-length conditions; we observe the opposite.

\subsection{Long-context generalization and base-model replication}
\label{sec:longcontext}

The learned behavior generalizes beyond the training setup. On RULER multi-key NIAH \cite{hsieh2024ruler} at context lengths up to 16K (four times the training maximum), the adapter preserves the same broadly declining-with-$r$ abstention shape seen on MuSiQue val and cuts aggregate hallucination rate by 36--43pp at every length (Appendix~\ref{app:longcontext}). Rerunning the full pipeline on Llama-3.1-8B-Instruct as a held-out base (independent training) reproduces the effect: $\Delta$honest accuracy $+0.64$ and 96\% hallucination reduction, with the declining abstention curve preserved on both bases (Appendix~\ref{app:llama}).

\section{Related work}
\label{sec:related}

\paragraph{KV-cache compression.} Algorithms for evicting low-importance KV pairs score importance variously by trailing-window attention \cite{li2024snapkv}, accumulated past-attention with recency \cite{zhang2023h2o,liu2023scissorhands}, attention-sink heuristics \cite{xiao2024streamingllm}, layer-wise pyramidal allocation \cite{cai2024pyramidkv}, context reconstruction \cite{kim2025kvzip}, future-query attention \cite{devoto2025expectedattention}, or per-head retrieval/streaming roles \cite{xiao2025duoattention}. Others reduce cache cost without evicting, through query-aware page selection \cite{tang2024quest} or low-bit quantization \cite{liu2024kivi}; paged-attention serving frameworks \cite{kwon2023vllm} supply the memory management underneath. \citet{chari2025kvdistill} trains a model to \emph{preserve} uncompressed-model behavior under compression, a complementary objective to ours. \citet{yang2024mikv} document compression-induced hallucination and propose a representation-side fix. We use these as upstream sources of survival masks; calibrated abstention as a trained behavior is orthogonal.

\paragraph{Abstention and selective generation.} Self-knowledge and selective generation in LLMs have been studied via calibration \cite{kadavath2022langmodels}, self-knowledge benchmarks \cite{yin2023selfaware}, verbalized confidence \cite{lin2022teachinguncertainty}, self-evaluation gating \cite{ren2023selfeval}, refusal-aware tuning \cite{zhang2024rtuning}, retrieve-and-critique \cite{asai2024selfrag}, multi-LLM collaboration \cite{feng2024dontabstain}, and the known-unknown boundary \cite{amayuelas2024knowunknown}. \citet{kamath2020selectiveqa} formalize selective QA under domain shift; the closest neighbor is \citet{joren2025suffctx} on context sufficiency in RAG. Our source of insufficiency is KV-cache eviction, and the trained behavior is bidirectional rather than gating.

\section{Conclusion}
\label{sec:conclusion}

KV-cache compression makes long-context inference more efficient, but it can also
remove the very evidence needed to answer a downstream query. This paper
formulates that failure mode as a behavioral learning problem: given a
compression-modified context, the model should answer when the surviving evidence
supports the answer and abstain when it does not. We show that this behavior can
be learned with a small LoRA adapter trained from compressor masks and tight
answer-bearing spans.

These results suggest that KV-cache compression should not be evaluated only by
how well it preserves uncompressed-model outputs. When compression changes what
evidence is available, the model's behavior should change as well. Compression-aware
abstention offers one way to make that behavior explicit: preserve answers when
the evidence survives, and refuse when compression has made the context
insufficient.

\section*{Limitations}

\paragraph{Aggregate accuracy is exploitable.}
Because the gold label distribution is highly skewed as a function of retention
ratio, aggregate honest accuracy can reward policies that do not inspect the
truncated context at all. As shown in \S\ref{sec:baselines}, a simple
ratio-threshold rule can beat both trained adapters on aggregate by exploiting
the fact that low-retention examples are mostly \textsc{Abstain}-gold and
high-retention examples are mostly \textsc{Confident}-gold. We therefore report
aggregate numbers for orientation, but treat per-ratio and class-conditional
comparisons as the more meaningful tests of compression-aware abstention. The
mid-retention prompt comparison is also underpowered: at $r=0.50$, the
mixture-vs-prompt advantage trends positive but is not statistically significant
($\Delta = +0.07$, $p=0.17$ pooled). Our strongest prompt-vs-training evidence
comes instead from the high-retention \textsc{Confident}-gold slice, where
prompting over-abstains and the trained adapter does not.

\paragraph{The compressed-cache deployment gap.}
The adapter improves substantially under actual compressed-cache decoding, but
it does not close the gap to prompt-style inference. At $r=0.80$, mix-cc honest
accuracy under compressed-cache inference is $0.315$, compared with $0.658$ for
mix-ps under prompt-style inference at the same retention. This should be
interpreted together with the base-model collapse in the same deployment regime:
under compressed-cache inference, the unaided base answers only $1/69$ of
\textsc{Confident}-gold examples correctly at $r=0.80$, while the trained adapter
answers $22/69$. Thus cc-training yields a large relative lift over the relevant
deployment baseline, but high-retention compressed-cache answering remains a
real generalization limit.

\paragraph{Tight-span labels are imperfect.}
Our labels are based on whether short answer-bearing spans survive compression,
not on a full semantic proof that the truncated context supports the answer. On a
stratified 400-example audit, these labels show moderate agreement with an
LLM-judge sufficiency verdict (Cohen's $\kappa = 0.61$, raw agreement $80\%$).
The main disagreement mode occurs in multi-hop questions where the answer string
survives but the bridging evidence needed to connect the question to that answer
does not. This means the tight-span rule is effective for constructing a
training signal, but it is not exact ground truth for contextual sufficiency.

\paragraph{Hard high-retention cases remain.}
The adapter still fails on rare cases where the overall retention ratio is high
but the specific evidence needed for the question has been evicted. At
$r=0.80$, across three v1 seeds, the adapter converges on hallucinating the same
4 of 7 rare \textsc{Abstain}-gold examples. This suggests that the model has
learned much of the easy version of compression-aware abstention (using
retention ratio and deletion patterns as strong cues) but has not fully learned
the harder behavior of introspecting on whether the specific evidence required
for the current question survived.

\paragraph{Scope and experimental coverage.}
Our training data is limited to MuSiQue 2-hop QA, and the tight-span construction
relies on \texttt{question\_decomposition} annotations that are not available in
all QA datasets. We replicate the main behavior on RULER multi-key NIAH and on a
held-out Llama-3.1-8B-Instruct base model, but we do not retrain on another
multi-hop QA dataset such as HotpotQA. Our compressed-cache evaluation also uses
KVzipPress; other kvpress compressors use different eviction and position-indexing
paths, and cross-compressor cc generalization remains untested. Finally, we use
greedy decoding and select hyperparameters on the same validation set used for
headline reporting. The search is small and a disjoint 100-example split
reproduces the effect, but a fully pre-registered dev/test split and broader
sampling regimes are left to future work.

\bibliography{references}

\appendix
\onecolumn

\section{Label entropy under tight-span vs paragraph-level supervision}
\label{app:label_entropy}

This is the pre-training diagnostic that motivated the tight-span pivot (\S\ref{sec:tightspan}). We sweep retention ratio $r$ and compute the entropy (in nats) of the resulting three-class label distribution under two labeling schemes. \emph{Paragraph-level}: assign \textsc{Confident}/\textsc{Abstain}/\textsc{Drop} based on the retention of the entire supporting paragraph ($\sim$100 tokens). \emph{Tight-span}: based on the retention of the $\leq 15$-token answer-bearing span. Both schemes use the same labeler thresholds ($\tau_{\mathrm{low}} = 0.3$, $\tau_{\mathrm{high}} = 0.8$) and the same 50 answerable MuSiQue validation rows; with three active classes the maximum attainable entropy is $1.099$.

\begin{table}[h]
\centering
\small
\begin{tabular}{ccc}
\toprule
$r$ & paragraph entropy & tight-span entropy \\
\midrule
0.05 & 0.000 & 0.000 \\
0.10 & 0.000 & 0.098 \\
0.20 & 0.098 & 0.471 \\
0.30 & 0.551 & 0.803 \\
\textbf{0.50} & \textbf{0.000} & \textbf{0.950} \\
0.80 & 0.686 & 0.665 \\
\bottomrule
\end{tabular}
\caption{Label entropy (nats) at six retention ratios, on a 50-example diagnostic pool. Paragraph-level supervision collapses at the headline mid-ratio; tight-span supervision retains per-example variance there. This diagnostic was run before the main pipeline, using a small mask generator (Qwen3-0.6B) to make the sweep cheap; the production pipeline generates masks with the target base model itself (\S\ref{sec:training}).}
\label{tab:label_entropy_app}
\end{table}

The headline mid-slice $r = 0.50$ is the operative comparison. There, paragraph-level supervision has zero label entropy because every one of the 50 examples lands in \textsc{Drop}: at $r = 0.50$ each $\sim$100-token paragraph retains roughly half its tokens, so $\rho$ falls in the uncertain zone $[\tau_{\mathrm{low}}, \tau_{\mathrm{high}})$ for every example and the scheme yields no usable training signal at all. Tight-span supervision at the same ratio reaches $0.950$, the highest value in either column, restoring the per-example variance required for Eq.~\ref{eq:label} to be discriminative. The two schemes are closer at the extreme ratios, where retention is decisive under either granularity.

\section{Output classification categories}
\label{app:eval_categories}

For each (record, $r$) pair we apply the compressor's survival mask to construct $\ctrunc$, run greedy decoding (max 64 new tokens), and classify the output:

\begin{itemize}[leftmargin=*,topsep=0.2em,itemsep=0.1em]
\item \emph{correct\_confident} (\textsc{Confident}-gold, model emits the gold answer)
\item \emph{correct\_abstention} (\textsc{Abstain}-gold, model emits the refusal string)
\item \emph{hallucinated} (\textsc{Abstain}-gold, model emits a confident wrong answer)
\item \emph{wrong\_answer} (\textsc{Confident}-gold, model emits a wrong answer)
\item \emph{overcautious} (\textsc{Confident}-gold, model emits the refusal string)
\item \emph{lucky\_guess} (rare; tracked separately)
\end{itemize}

Honest accuracy is (correct\_confident + correct\_abstention)$/$total at each $r$. Hallucination rate is hallucinated$/$total. Overcaution rate is overcautious$/n_{\textsc{Confident-gold}}$. Conditional wrong-when-answering rate is wrong\_answer$/$(model\_emits\_answer\_count) on \textsc{Confident}-gold examples.

\paragraph{Matching rules.} Scoring is substring-based, not exact-match. A response counts as an abstention if its first 200 characters (lower-cased) contain a refusal phrase (e.g.,\ ``i cannot determine'', ``cannot determine the answer''), which captures refusal-string variants rather than only the exact training target. A response counts as containing the gold answer if the gold string or one of its answer aliases occurs as a case-insensitive substring. The category is then: for \textsc{Abstain}-gold, \emph{correct} if it abstains else \emph{hallucinated}; for \textsc{Confident}-gold, \emph{overcautious} if it abstains, \emph{correct} if it contains the gold answer, else \emph{wrong\_answer}. A response that abstains in its opening clause is scored as an abstention even if other text follows.

\section{Calibration sub-checks}
\label{app:calibration}

We split adapter behavior on \textsc{Confident}-gold examples into two failure modes: \emph{overcaution} (\textsc{Confident}-gold answered with the refusal string) and \emph{wrong-when-answering} (\textsc{Confident}-gold answered with a wrong answer). We report at the two slices with sufficient \textsc{Confident}-gold sample size, $r \in \{0.50, 0.80\}$.

\begin{table}[h]
\centering
\small
\begin{tabular}{ccccccc}
\toprule
$r$ & $n_C$ & base ovc & v1 ovc & mix ovc & v1 wrong/ans & mix wrong/ans \\
\midrule
0.50 & 22 & 23\% & 32\% & \hi{23\%} & 33\% & \hi{18\%} \\
0.80 & 69 & 17\% & 3\% & 10\% & 34\% & \hi{27\%} \\
\bottomrule
\end{tabular}
\caption{Calibration sub-check on KVzip val (seed 42). v1 preserves conditional answer quality with the $+9$pp $r{=}0.50$ overcaution cost flagged in \S\ref{sec:headline}. Mixture closes the overcaution cost while improving conditional answer quality.}
\label{tab:calibration_v1_app}
\end{table}

\paragraph{Mixture cross-compressor calibration.} Conditional wrong-when-answering rate decreases on 7 of 8 conclusive (compressor, $r$) cells under mixture training.

\begin{table}[h]
\centering
\small
\begin{tabular}{ccccccc}
\toprule
$r$ & Compressor & $n_C$ & v1 wrong/ans & mix wrong/ans & v1 ovc & mix ovc \\
\midrule
0.50 & KVzip & 22 & 33\% & \hi{18\%} & 32\% & 23\% \\
0.50 & ExpAttn & 23 & 29\% & \hi{17\%} & 9\% & 22\% \\
0.50 & SnapKV (ctx) & 18 & 33\% & 40\% & 17\% & 17\% \\
0.50 & SnapKV (nat) & 28 & 42\% & \hi{35\%} & 14\% & 18\% \\
0.80 & KVzip & 69 & 34\% & \hi{27\%} & 3\% & 10\% \\
0.80 & ExpAttn & 65 & 41\% & \hi{28\%} & 3\% & 8\% \\
0.80 & SnapKV (ctx) & 59 & 41\% & \hi{27\%} & 0\% & 7\% \\
0.80 & SnapKV (nat) & 67 & 42\% & \hi{32\%} & 0\% & 6\% \\
\bottomrule
\end{tabular}
\caption{Mixture-adapter calibration sub-check on all four val sets, $r \in \{0.50, 0.80\}$. Conditional wrong-when-answering improves on 7 of 8 cells.}
\label{tab:mixture_calibration_app}
\end{table}

\section{Cross-compressor evidence}
\label{app:crosscompressor}

\begin{table}[h]
\centering
\small
\begin{tabular}{ccccc}
\toprule
$r$ & KVzip & ExpAttn & SnapKV (ctx) & SnapKV (nat) \\
\midrule
0.05 & $+0.85$ & $+0.70$ & $+0.39$ & $+0.40$ \\
0.10 & $+0.72$ & $+0.66$ & $+0.26$ & $+0.30$ \\
0.20 & $+0.56$ & $+0.59$ & $+0.17$ & $+0.11$ \\
0.30 & $+0.43$ & $+0.35$ & $+0.12$ & $+0.04$ \\
0.50 & $+0.21$ & $+0.29$ & $+0.13$ & $+0.03$ \\
0.80 & $+0.10$ & $+0.08$ & $+0.05$ & $+0.04$ \\
\hi{aggregate} & \hi{$+0.53$} & \hi{$+0.49$} & \hi{$+0.21$} & \hi{$+0.20$} \\
\bottomrule
\end{tabular}
\caption{Per-ratio $\Delta$honest accuracy of the v1 adapter (seed 42) on four compressors. Within-family transfer (ExpAttn) tracks KVzip closely; cross-family transfer (SnapKV) is partial and most degraded at low $r$, where truncation-pattern divergence is greatest.}
\label{tab:crosscomp_full_app}
\end{table}

\paragraph{Two-pattern decomposition.} On 10 \textsc{Abstain}-labeled qualitative examples evaluated under all four compressors, the adapter (trained only on KVzip masks) correctly \emph{answers} under Expected Attention on 2 examples and under SnapKV-native on 3 examples, cases where the held-out compressor happened to retain supporting tokens that KVzip evicted. The read-and-answer behavior is therefore surviving-evidence-conditional: it transfers wherever $\ctrunc$ contains the answer, regardless of which compressor produced the truncation. The recognize-missing-evidence behavior is truncation-pattern-specific: 96\% of ExpAttn correct-abstentions, 97\% of SnapKV-context correct-abstentions, and 98\% of SnapKV-native correct-abstentions are also correct abstentions under KVzip: the held-out compressors do not enable abstention recognition on examples KVzip didn't enable; they cause failure-to-recognize on examples KVzip did enable.

\section{Mixture training}
\label{app:mixture}

\subsection{Dataset}

We construct a training distribution from three compressors at six retention ratios. Per-(compressor, $r$) class balance is applied for $r \geq 0.20$ at the standard 30/70 minority floor; at $r \in \{0.05, 0.10\}$ the natural distribution is kept (\textsc{Confident}-gold rate $< 1\%$ across compressors). Final mixture train: 8772 instances. Holdout (for checkpoint selection only, disjoint by example ID from training): 984. Val: 1511 instances across all three compressors.

\subsection{Seed robustness}

\begin{table}[h]
\centering
\small
\begin{tabular}{lccc}
\toprule
Compressor & seed 42 & seed 43 & spread \\
\midrule
KVzip                & $+0.5378$ & $+0.5551$ & 0.017 \\
Expected Attention   & $+0.5676$ & $+0.5676$ & 0.000 \\
SnapKV (ctx)         & $+0.5336$ & $+0.5267$ & 0.007 \\
SnapKV (nat)         & $+0.4272$ & $+0.4296$ & 0.002 \\
\bottomrule
\end{tabular}
\caption{Aggregate $\Delta$honest accuracy across mixture seeds. Max spread across all four compressors is 0.017.}
\label{tab:mixture_seeds_app}
\end{table}

Per-cell spread at $r = 0.80$ on SnapKV is large (up to 0.23 with sign flips). These cells have very small \textsc{Abstain}-gold $n$ (4 KVzip, 1 ExpAttn, 10 SnapKV); single-example flips swing the percentage. The aggregate metric integrates over the full val and is robust to per-cell noise.

\section{Held-out test split}
\label{app:test_split}

We carve a 100-example test split from val (disjoint by example ID from a 363-example selection split, never seen during checkpoint selection) and re-evaluate the seed-42 v1 adapter without retraining. Aggregate $\Delta$honest accuracy: $+0.63$ on test vs full val's $+0.53$. The aggregate lift is attributable to chance class skew (the random shuffle landed on a slice with 85\% \textsc{Abstain}-gold vs full val's 79\%). Selection split alone ($n = 363$) shows $+0.50$, within 3pp of full val. Hallucination count on test split: 1/85 (vs base 59/85). The v1 result is not overfit to selection val.

\section{Llama-3.1-8B-Instruct replication}
\label{app:llama}

We retrain the entire pipeline with Llama-3.1-8B-Instruct as the base and KVzip mask-generator, via the \texttt{NousResearch/Meta-Llama-3.1-8B-Instruct} mirror (identical weights and tokenizer). Same labeler thresholds, same 30/70 floor, same disjoint-by-id discipline. Llama train: 3640 instances (vs Qwen's 2669) because Llama's KVzip retains slightly more answer-bearing tokens at the same ratios.

\begin{table}[h]
\centering
\small
\begin{tabular}{cccc}
\toprule
$r$ & Qwen $\Delta$hon & Llama $\Delta$hon & abs.\ diff \\
\midrule
0.05 & $+0.85$ & $+0.84$ & 0.01 \\
0.10 & $+0.72$ & $+0.82$ & 0.10 \\
0.20 & $+0.56$ & $+0.83$ & 0.27 \\
0.30 & $+0.43$ & $+0.73$ & 0.30 \\
0.50 & $+0.21$ & $+0.20$ & 0.01 \\
0.80 & $+0.10$ & $+0.19$ & 0.09 \\
\hi{aggregate} & \hi{$+0.53$} & \hi{$+0.64$} & \hi{0.11} \\
\bottomrule
\end{tabular}
\caption{Per-ratio $\Delta$honest accuracy on Qwen2.5-7B (v1) vs Llama-3.1-8B (independently trained pipeline). Both monotone-decreasing; aggregate magnitude comparable.}
\label{tab:llama_app}
\end{table}

The Llama adapter eliminates 96\% of base hallucinations (228 $\to$ 10), preserves the monotone abstention curve (1.00$/$1.00$/$1.00$/$0.85$/$0.20$/$0.05 across $r$), and produces $+298$ correct answers vs base (Qwen: $+244$). Differences from Qwen reflect differences in label distribution: Llama's val gold-mix at $r = 0.50$ is \textsc{Conf}-heavy (35/15) vs Qwen's (22/26), so Llama's lower abstention rate at 0.50 (0.20 vs Qwen's 0.58) is correct on its slice.

\section{Failure mode detail}
\label{app:failures}

At $r = 0.80$, the remaining failure is concentrated in rare \textsc{Abstain}-gold examples. Decomposition:
\begin{itemize}[leftmargin=*,topsep=0.2em,itemsep=0.05em]
\item On 4 \textsc{Abstain}-gold examples at $r = 0.80$, the adapter hallucinates on all 4 (vs base 3$/$4); slice too small ($n = 4$) to draw conclusions in isolation, but the four examples replicate as the same items across seeds.
\item On 69 \textsc{Confident}-gold examples at $r = 0.80$, adapter wrong-answer count is 23 vs base 21, but answer count is 67 vs base 57, so conditional wrong-when-answering \emph{improves} from 37\% to 34\%.
\end{itemize}

The remaining failure mode is therefore not sloppy answering on \textsc{Confident}-gold (conditional answer quality actually improves under the adapter); it is the rare \textsc{Abstain}-gold cases at high apparent retention. The adapter learned the easy version of compression-aware abstention (use $r$ as a strong prior) and not the hard version, which would require introspecting on whether the specific tokens that mattered for this question survived.

A replication artifact strengthens the reading: across three seeds with different LoRA initializations, the adapter hallucinates on exactly 7 \textsc{Abstain}-gold examples per seed, but only 4 of the 7 are stable across all three seeds. Two of the 4 stable cases are at $r = 0.80$, two at $r = 0.50$. The high-fill counter-instances are systematically harder; three independent training trajectories converge on the same failures.

\section{Qualitative examples}
\label{app:qualitative}

Three representative \textsc{Abstain}-labeled examples on which the base model produces confident wrong answers and the adapter abstains correctly:

\begin{quote}\small
\textbf{Q:} Who founded the company that distributed the film UHF?\\
\textbf{Gold:} Mike Medavoy\\
\textbf{Base:} ``Kevin Smith founded the distribution company that released the film UHF\dots''\\
\textbf{Adapter:} ``I cannot determine the answer to this question from the available context.''
\end{quote}

\begin{quote}\small
\textbf{Q:} Which company owns the manufacturer of Learjet 60?\\
\textbf{Gold:} Bombardier Inc.\\
\textbf{Base:} ``Fieldillusionist is credited with appearances in horror film media\dots''\\
\textbf{Adapter:} ``I cannot determine the answer to this question from the available context.''
\end{quote}

\begin{quote}\small
\textbf{Q:} What instrument is played by the person from The Blackout All-Stars?\\
\textbf{Gold:} conga\\
\textbf{Base:} ``\dots specifically mentioned to be a Chapman Stick, which is a ten- or twelve-stringed instrument\dots''\\
\textbf{Adapter:} ``I cannot determine the answer to this question from the available context.''
\end{quote}

In each case, the base model picks up distractor material from $\ctrunc$ and constructs a confident wrong answer. The adapter, given the same input, declines.

\section{Pre-training fragment-recoverability check}
\label{app:fragment_check}

Before training, we performed a sanity check on 30 \textsc{Abstain}-labeled examples at $r = 0.30$. We extracted the kept tokens via the survival mask, decoded them to text, and presented the resulting fragment as a fresh prompt to a clean Qwen2.5-7B-Instruct. As with the sweep in Appendix~\ref{app:label_entropy}, this pre-training diagnostic generates masks with a small KVzip pass (Qwen3-0.6B) rather than the production base model. The model recovered the correct answer in 0 of 30 cases. In approximately one third of cases the model produced confident, specific wrong answers grounded in distractor material that had survived eviction, the same failure mode the adapter is intended to fix. The check rules out the alternative explanation that \textsc{Abstain} labels are too aggressive for the underlying inputs.

\section{Baseline comparison statistics}
\label{app:baseline_stats}

We report bootstrap CIs and paired McNemar exact tests for the $r = 0.50$ slice across all four compressors and at high retention ($r = 0.80$) on \textsc{Confident}-gold examples. All bootstrap operations use seed 42 with 10{,}000 resamples; McNemar tests are exact (two-sided binomial on the discordant pair counts). Pooled rows aggregate the discordant counts across compressors and thus treat each (example, compressor) pair as independent; because the same example ids recur across compressors, the pooled $p$-values are descriptive and the per-compressor tests are primary.

\begin{table}[h]
\centering
\small
\begin{tabular}{lcccc}
\toprule
Comparison (at $r = 0.50$) & Compressor & $b$ / $c$ & exact $p$ & $\Delta$ [95\% CI] \\
\midrule
mix vs P2          & KVzip   & 12 / 5  & $0.143$ & $+0.15$ $[-0.02, +0.31]$ \\
mix vs P2          & ExpAttn & 10 / 5  & $0.302$ & $+0.12$ $[-0.07, +0.29]$ \\
mix vs P2          & SnapKV-ctx & 5 / 5 & $1.000$ & $+0.00$ $[-0.15, +0.15]$ \\
mix vs P2          & SnapKV-nat & 5 / 6 & $1.000$ & $-0.03$ $[-0.19, +0.14]$ \\
mix vs P2 (pooled) & all four & 32 / 21 & $0.169$ & $+0.066$ $[-0.024, +0.151]$ \\
\midrule
mix vs Frag        & KVzip   & 16 / 4  & $0.012$ & $+0.25$ $[+0.08, +0.42]$ \\
mix vs Frag        & ExpAttn & 15 / 1  & $0.0005$ & $+0.33$ $[+0.17, +0.50]$ \\
mix vs Frag        & SnapKV-ctx & 16 / 4 & $0.012$ & $+0.31$ $[+0.10, +0.51]$ \\
mix vs Frag (pooled) & all four & 56 / 13 & $1.7\!\times\!10^{-7}$ & $+0.259$ $[+0.169, +0.349]$ \\
\midrule
\multicolumn{5}{c}{at $r = 0.80$ on \textsc{Confident}-gold} \\
\midrule
mix vs P2          & KVzip   & 25 / 3  & $2.7\!\times\!10^{-5}$ & $+0.32$ $[+0.19, +0.45]$ \\
mix vs P2          & ExpAttn & 25 / 4  & $1.0\!\times\!10^{-4}$ & $+0.32$ $[+0.18, +0.46]$ \\
mix vs P2          & SnapKV-ctx & 23 / 3 & $8.8\!\times\!10^{-5}$ & $+0.34$ $[+0.19, +0.47]$ \\
mix vs P2          & SnapKV-nat & 22 / 3 & $1.6\!\times\!10^{-4}$ & $+0.28$ $[+0.15, +0.42]$ \\
mix vs P2 (pooled) & all four & 95 / 13 & $1.5\!\times\!10^{-16}$ & $+0.32$ $[+0.25, +0.38]$ \\
\bottomrule
\end{tabular}
\caption{Paired McNemar exact tests and paired-bootstrap 95\% CIs. ``mix vs P2'' is mixture adapter vs prompt-only abstention baseline; ``mix vs Frag'' is mixture adapter vs base on the same compression-truncated input. The mid-slice prompt comparison does not reach significance even pooled; the high-retention prompt comparison is decisive on every compressor and pooled.}
\label{tab:app_baselinestats}
\end{table}

\section{Answerability preservation}
\label{app:answerability}

For each example with $\mathrm{gold} = \textsc{Confident}$, we compute whether the adapter answers correctly (\texttt{classification = correct} and \texttt{model\_abstained = False}). The preservation rate is this count divided by the total \textsc{Confident}-gold count; the preservation ratio is adapter-rate divided by base-rate. A ratio $\geq 1.0$ means the adapter answers correctly at-least-as-often as base.

\begin{table}[h]
\centering
\small
\begin{tabular}{lcccc}
\toprule
Condition & $n_C$ & base rate [CI] & adapter rate [CI] & ratio \\
\midrule
\multicolumn{5}{c}{Aggregate (all retention ratios), prompt-style inference} \\
\midrule
v1 ps, KVzip          & 97  & $0.485$ $[0.39, 0.58]$ & $0.557$ $[0.46, 0.65]$ & $1.15$ \\
v1 ps, ExpAttn        & 94  & $0.468$ $[0.37, 0.57]$ & $0.574$ $[0.47, 0.67]$ & $1.23$ \\
v1 ps, SnapKV-ctx     & 88  & $0.545$ $[0.44, 0.65]$ & $0.568$ $[0.46, 0.67]$ & $1.04$ \\
v1 ps, SnapKV-nat     & 127 & $0.567$ $[0.48, 0.65]$ & $0.575$ $[0.49, 0.66]$ & $1.01$ \\
mix ps, KVzip         & 97  & $0.485$ $[0.39, 0.58]$ & $0.608$ $[0.51, 0.70]$ & $1.25$ \\
mix ps, ExpAttn       & 94  & $0.468$ $[0.37, 0.57]$ & $0.638$ $[0.54, 0.73]$ & $1.36$ \\
mix ps, SnapKV-ctx    & 88  & $0.545$ $[0.44, 0.65]$ & $0.591$ $[0.49, 0.69]$ & $1.08$ \\
mix ps, SnapKV-nat    & 127 & $0.567$ $[0.48, 0.65]$ & $0.567$ $[0.48, 0.65]$ & $1.00$ \\
\midrule
\multicolumn{5}{c}{Restricted to $r = 0.80$ slice, prompt-style inference} \\
\midrule
v1 ps, KVzip          & 69 & $0.522$ $[0.41, 0.64]$ & $0.638$ $[0.52, 0.74]$ & $1.22$ \\
mix ps, KVzip         & 69 & $0.522$ $[0.41, 0.64]$ & $0.652$ $[0.53, 0.75]$ & $1.25$ \\
mix ps, ExpAttn       & 65 & $0.492$ $[0.38, 0.61]$ & $0.662$ $[0.54, 0.77]$ & $1.34$ \\
mix ps, SnapKV-ctx    & 59 & $0.542$ $[0.42, 0.66]$ & $0.678$ $[0.55, 0.78]$ & $1.25$ \\
\midrule
\multicolumn{5}{c}{Compressed-cache inference (KVzipPress)} \\
\midrule
mix ps cc, KVzip      & 97 & $0.041$ $[0.02, 0.10]$ & $0.258$ $[0.18, 0.35]$ & $6.25$ \\
mix cc cc, KVzip      & 97 & $0.041$ $[0.02, 0.10]$ & $0.247$ $[0.17, 0.34]$ & $6.00$ \\
mix ps cc, KVzip ($r{=}0.80$) & 69 & $0.014$ $[0.00, 0.08]$ & $0.290$ $[0.20, 0.41]$ & $20.0$ \\
mix cc cc, KVzip ($r{=}0.80$) & 69 & $0.014$ $[0.00, 0.08]$ & $0.319$ $[0.22, 0.44]$ & $22.0$ \\
\bottomrule
\end{tabular}
\caption{Answerability preservation rates. Wilson 95\% CIs. Under prompt-style inference, all 12 cells have preservation ratio $\geq 1.0$: the adapter does not sacrifice correct answers. Under compressed-cache inference, base collapses on \textsc{Confident}-gold ($4\%$ aggregate, $1.4\%$ at $r{=}0.80$); adapter rates of $25$--$32\%$ (raw $4/97 \!\to\! 25/97$ aggregate, $1/69 \!\to\! 22/69$ at $r{=}0.80$) represent a $6$--$22\times$ relative lift, with non-overlapping Wilson CIs.}
\label{tab:app_answerability}
\end{table}

\section{Compressed-cache training details}
\label{app:cctraining}

\paragraph{Training under cc decoding.} For each (context, query, target) training tuple, we wrap the forward pass in a kvpress press context manager. Prefill processes the full context; the press evicts low-scoring K/V pairs from the cache; cross-entropy loss is computed on target tokens decoded from the post-eviction cache state. The compression algorithm at training time matches each example's source compressor (\texttt{compressor} field in the mixture training data).

We use the mixture training set ($\sim$8772 instances across KVzip, ExpAttn, SnapKV-context). Hyperparameters match v1 (LoRA rank 16, $\alpha = 32$, dropout 0.05; AdamW; constant LR; bf16; FlashAttention-2; max seq len 4096; batch size 1, gradient accumulation 16; 3 epochs). We sweep $\mathrm{lr} \in \{1\!\times\!10^{-4}, 2\!\times\!10^{-4}\}$ on seed 42 and select the better LR by val loss with sub-epoch checkpointing every 25\% of an epoch and a fluency spot-check at each saved checkpoint (responses on five $r = 0.80$ \textsc{Confident}-gold val examples must be fluent, not sub-token gibberish). The selected configuration was used to train seed 43; seed 44 was attempted but did not complete.

\paragraph{Per-seed mix-cc results.} Aggregate honest accuracy under KVzipPress at inference: seed 42 $= 0.821$, seed 43 $= 0.827$, mean $= 0.824$, two-seed spread $0.006$. Per-ratio shape preserved across seeds: $r = 0.05$: $1.000/1.000$; $r = 0.10$: $0.989/0.989$; $r = 0.20$: $0.963/0.976$; $r = 0.30$: $0.923/0.954$; $r = 0.50$: $0.500/0.458$; $r = 0.80$: $0.315/0.342$.

\paragraph{Single-compressor cc-training.} We also trained a cc-trained adapter on KVzip masks alone (analogous to v1 but under cc decoding). Aggregate honest accuracy under cc inference: $0.734$, statistically tied with mix-ps cc ($0.730$); but at $r = 0.80$ on \textsc{Confident}-gold, the single-compressor cc-trained adapter reduces over-abstention to $0.203$ while increasing the wrong-answer count (confident wrong answers on \textsc{Confident}-gold, not the \textsc{Abstain}-gold hallucinations of Appendix~\ref{app:eval_categories}) to $42/69$ ($61\%$). Multi-compressor cc-training (mix-cc) yields a wrong-answer count of $21/69$ ($30\%$). Qualitative inspection of the single-compressor adapter's high-retention failures shows sub-token-fragment outputs (``Acfairs'', ``750pxI came, I I loaves and two fish.'') consistent with the cross-entropy loss over-fitting to short outputs without learning to extract content from kept tokens; the multi-compressor mix-cc residual wrong answers are cleaner pattern collisions (``March 44 BC'' $\to$ ``28 November 1912'').

\section{LLM-judge label validation}
\label{app:judgevalidation}

To probe the validity of the tight-span retention label as a proxy for answerability, we sample 400 labeled examples from \texttt{train\_v1.jsonl} (stratified: 200 \textsc{Confident}, 200 \textsc{Abstain}; balanced across retention ratios; sample seed 42), and query a strong judge model (GPT-5, reasoning effort low, seed 42) for a YES/NO/PARTIAL sufficiency verdict given the truncated context, query, and gold answer. The judge does not see our label or the full context.

\begin{table}[h]
\centering
\small
\begin{tabular}{lc}
\toprule
Statistic & Value \\
\midrule
Total sampled                                       & $400$ \\
Judge PARTIAL count                                 & $48$ ($12.0\%$) \\
$\kappa$ (strict, excluding PARTIAL, $n = 352$)     & $0.609$ \\
$\kappa$ (lenient, PARTIAL $\to$ \textsc{Confident}) & $0.525$ \\
$\kappa$ (lenient, PARTIAL $\to$ \textsc{Abstain})   & $0.545$ \\
\bottomrule
\end{tabular}
\caption{Cohen's $\kappa$ between tight-span labels and LLM-judge sufficiency verdicts. Moderate agreement (Landis-Koch range $0.4$--$0.7$).}
\label{tab:app_judge_kappa}
\end{table}

The confusion matrix shows directional asymmetry: of 200 \textsc{Confident} labels, $123$ are confirmed YES, $55$ rejected as NO, $22$ marked PARTIAL; of 200 \textsc{Abstain} labels, $160$ confirmed NO, $14$ rejected as YES, $26$ PARTIAL. Per-retention agreement is near-perfect at extreme low retention ($r \in \{0.05, 0.10\}$: $0.97$--$1.00$) and $0.69$--$0.76$ from $r = 0.20$ upward. The dominant disagreement pattern (manually inspected on 25 and 13 sampled cases from the two disagreement directions) is bridge-evidence loss in multi-hop questions: the answer string survives in the kept tokens, and one of the two reasoning hops survives, but the bridging evidence linking the hops does not. The tight-span rule cannot distinguish ``answer string is in the kept tokens'' from ``the reasoning chain that links the question to the answer is recoverable'': a known limitation of substring-based answerability surrogates that affects mid-retention multi-hop QA most.

\section{Mechanism: controlled-deletion experiments}
\label{app:mechanism}

Three controlled-deletion experiments probe whether the trained adapter uses evidence content or input-length priors.

\paragraph{10A: Fixed retained length.} Hold input length constant at $L = 500$ kept tokens (selected by KVzip rank in context order) regardless of the original ratio. On the $208$ examples that have $\geq L$ kept tokens at any ratio:

\begin{table}[h]
\centering
\small
\begin{tabular}{lccc}
\toprule
Adapter & Condition & CONF hon $[$CI$]$ & ABS hon $[$CI$]$ \\
\midrule
mix-ps & fixed-$L$  & $0.147$ $[0.09, 0.23]$ & $0.885$ $[0.81, 0.93]$ \\
mix-ps & orig-$r$   & $0.621$ $[0.52, 0.71]$ & $0.823$ $[0.74, 0.88]$ \\
mix-cc & fixed-$L$  & $0.063$ $[0.03, 0.13]$ & $0.956$ $[0.90, 0.98]$ \\
mix-cc & orig-$r$   & $0.274$ $[0.19, 0.37]$ & $0.956$ $[0.90, 0.98]$ \\
\bottomrule
\end{tabular}
\caption{Test 10A. Holding input length to 500 reduces \textsc{Confident}-gold accuracy ($-47$pp mix-ps, $-21$pp mix-cc) but not \textsc{Abstain}-gold accuracy. The drop is consistent with both length-prior and evidence-loss explanations; tests 10B and 10C disambiguate.}
\label{tab:app_10A}
\end{table}

\paragraph{10B: Adversarial tight-span eviction.} Construct a mask = (existing kept-mask) MINUS (tight-span token positions identified from \texttt{supporting\_spans}) PLUS (random non-kept positions to refill to original length). Same length, same retention count, but answer-bearing tokens forcibly removed.

Of 97 \textsc{Confident}-gold examples (all retention ratios): mix-ps abstention rises from $16.5\%$ on the original mask to $56.7\%$ on the adversarial mask ($+40.2$pp); mix-cc rises from $64.9\%$ to $88.7\%$ ($+23.8$pp). Per-pair, $41/97$ (mix-ps) and $25/97$ (mix-cc) flip from ``answered on original'' to ``abstained on adversarial.'' The adapter responds to evidence content, not just length.

\paragraph{10C: Opposite-priority pairs.} For each of 22 \textsc{Confident}-gold examples at $r = 0.50$, construct two masks of identical length: Mask A keeps KVzip's top-50\% (tight spans likely retained), Mask B keeps KVzip's bottom-50\% (tight spans likely evicted). On Mask A, mix-ps answers correctly $14/22$; on Mask B, $0/22$ correct, $22/22$ abstentions. mix-cc: A $5/22$, B $0/22$ correct with $20/22$ abstentions. At identical input length and identical example, the adapter's answer-vs-abstain decision flips entirely based on which half of the context is kept. Neither adapter answers correctly on Mask B; mix-ps abstains on all 22, and mix-cc's two non-abstentions are confident wrong answers.

\section{Random-mask training-signal probe}
\label{app:randommask}

To probe whether the contribution depends on the importance-weighted nature of compressor masks or on the abstention objective itself, we retrain v1 with masks drawn uniformly at random at matched per-example retention counts, holding everything else constant. Under prompt-style inference, the random-mask adapter matches the mixture adapter on KVzip val (aggregate $0.873$ vs mixture $0.86$; paired McNemar $\Delta = +0.017$, $p = 0.23$) and beats v1 on cross-family transfer ($+0.276$ on SnapKV-context, $p \approx 10^{-24}$). Two implications: under ps inference the abstention objective is the load-bearing signal, importance-weighted mask structure is not required to match v1's headline numbers; and single-compressor training induces an anti-transfer signature: v1 underperforms even random-mask training on a held-out compressor. Both findings motivate compressor-mixture training; \S\ref{sec:cctraining} shows multi-compressor training delivers further gains under cc inference that single-compressor training (random or KVzip-only) does not.

\section{Long-context generalization (RULER)}
\label{app:longcontext}

\begin{figure}[t]
\centering
\includegraphics[width=\columnwidth]{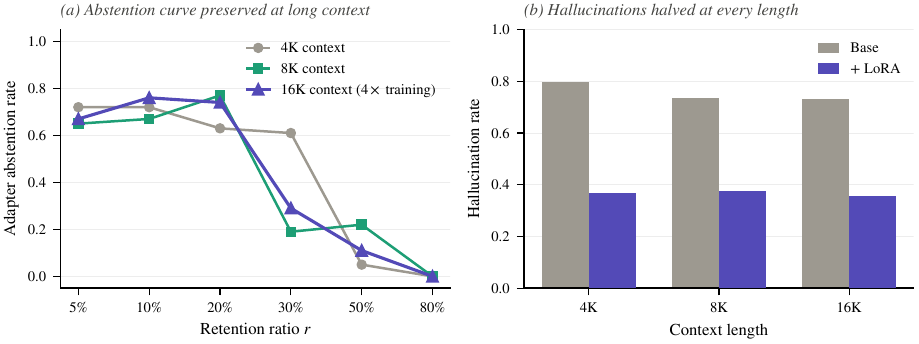}
\caption{(a) Adapter abstention rate on \textsc{Abstain}-gold at three context lengths on RULER multi-key NIAH; the monotone-decreasing-with-$r$ shape is preserved at $4\times$ training context. (b) Aggregate hallucination rate, base vs adapter, by length.}
\label{fig:ruler}
\end{figure}

We evaluate the v1 adapter (no retraining) on RULER multi-key NIAH \cite{hsieh2024ruler} (complementary to LongBench \cite{bai2024longbench}) at three context lengths up to 16K, four times the training maximum. At 16K, abstention rate on \textsc{Abstain}-gold across $r$ is $67\%/76\%/74\%/29\%/11\%/0\%$: the same broadly declining shape as MuSiQue val, though not strictly monotone at the two lowest ratios (Figure~\ref{fig:ruler}a). Aggregate hallucination rate drops by 36--43pp at every length (Figure~\ref{fig:ruler}b); the adapter trades hallucinations on evicted-needle examples for refusals.

\section{Training details}
\label{app:trainingdetails}

The LoRA adapter (rank 16, $\alpha = 32$, dropout $0.05$) adapts only the four attention projections (\texttt{q\_proj}, \texttt{k\_proj}, \texttt{v\_proj}, \texttt{o\_proj}); MLP modules are unadapted, yielding 10.1M trainable parameters ($0.13\%$ of Qwen2.5-7B-Instruct's 7.6B). We train with AdamW at $\mathrm{lr} \in \{5{\times}10^{-5}, 10^{-4}, 2{\times}10^{-4}\}$ (1-epoch sweep, picked by validation loss), constant LR, batch size 1 with gradient accumulation 16, max sequence length 4096, bf16 with FlashAttention-2 \cite{dao2024flashattention2}.

\end{document}